\documentclass{article}

\usepackage{microtype}
\usepackage{amsmath, amssymb, amsthm, amsfonts}
\usepackage{graphicx}
\usepackage{subfigure}
\usepackage[dvipsnames]{xcolor}
\usepackage{booktabs, pifont} 

\usepackage[accepted]{tccmlicml2021}

\icmltitlerunning{ANP-BBO for Calibrating Physics-Informed Digital Twins}

\newcommand{\ys}{y^\star_{0:T}}
\newcommand{\mt}{\mathcal M_T(\theta)}
\newcommand{\mtstar}{\mathcal M_T(\theta^\star)}
\newcommand{\y}{y_{0:T}}

\begin{document}
\setlength{\abovedisplayskip}{4pt}
\setlength{\belowdisplayskip}{4pt}

\twocolumn[
\icmltitle{ANP-BBO: Attentive Neural Processes and Batch Bayesian Optimization \\ for Scalable Calibration of  Physics-Informed Digital Twins}

\begin{icmlauthorlist}
\icmlauthor{Ankush Chakrabarty}{merl}
\icmlauthor{Gordon Wichern}{merl}
\icmlauthor{Christopher R. Laughman}{merl}
\end{icmlauthorlist}

\icmlaffiliation{merl}{Mitsubishi Electric Research Laboratories (MERL), Cambridge, MA, USA.}

\icmlcorrespondingauthor{A. Chakrabarty}{achakrabarty@ieee.org}

\icmlkeywords{Neural processes, Bayesian optimization, attention}

\vskip 0.3in
]

\printAffiliationsAndNotice{} 

\begin{abstract}
Physics-informed dynamical system models form critical components of digital twins of the built environment. These digital twins enable the design of energy-efficient infrastructure, but must be properly calibrated to accurately reflect system behavior for downstream prediction and analysis. Dynamical system models of modern buildings are typically described by a large number of parameters and incur significant computational expenditure during simulations. To handle large-scale calibration of digital twins without exorbitant simulations, we propose ANP-BBO: a scalable and parallelizable  batch-wise Bayesian optimization (BBO) methodology that leverages attentive neural processes (ANPs).
\end{abstract}

\section{Motivation}
Buildings account for nearly 40\% of global electricity use (over 70\% in the U.S.) and at least one third of $\mathrm{CO_2}$ emissions, while space cooling specifically plays a prominent role as it represents more than 70\% of peak residential electricity demand to cope with extreme weather. Forecasts indicate that the demand for space cooling will continue rapid growth, with the energy consumed by these applications projected to triple between 2016 and 2050~\cite{IEA2018}. Current efforts to reduce the climate-related impact of this energy consumption are focused on the creation of grid-interactive buildings, which coordinate the dynamic behavior of buildings with electrical grid behavior that is dominated by time-varying distributed energy resources~\cite{DOE2021}. As the design and control of these buildings represent a significant change in how buildings are operated, new models that accurately predict their experimentally-observed dynamics, the so-called building `digital twins', are crucial to developing these next-generation systems.  

Building and heating, ventilation, and cooling (HVAC) digital twins need to be calibrated to operational data to accurately replicate the observed behavior of the physical system. Physics-informed dynamical models have a number of advantages in digital twin applications, as they have good predictive/extrapolation properties, their parameters are interpretable by domain experts, and they can  be built using information that is measured or archived. Unfortunately, these advantages are often accompanied by nonlinear behavior and numerical stiffness that make simulation sluggish, and the models often comprise translucent/opaque components  for privacy or proprietary information security. The ensuing calibration problem therefore tends to be black-box  and large, because modern digital twins often contain hundreds or thousands of parameters to be calibrated. Machine learning has been identified as a key technology in optimizing building models~\cite{rolnick2019tackling}.

This calibration problem can be abstracted by considering a predictive simulation model 
\begin{equation}\label{eq:model_map_mt}
\y=\mt,
\end{equation}  
where $\theta\!\!\in\!\!\Theta\!\!\subset\!\!\mathbb R^{n_\theta}$ denotes the constant parameters used to parameterize the building and HVAC dynamics. A search domain of parameters $\Theta$ is assumed to be available, and we assume $\Theta$ is a box in $n_\theta$-space defined by bounded intervals. 
The output vector $\y\in\mathbb R^{n_y\times T}$ denotes the outputs that have been measured using the real building sensors over a time-span $[0, T]$. We do not make any assumptions on the underlying mathematical structure of the model $\mt$, except that it has been designed based on building and HVAC physics, implying that the parameters and outputs are interpretable physical quantities. Simulating $\mt$ forward with a set of parameters $\theta\in\Theta$ yields a vector of outputs
$\y := \begin{bmatrix}
y_0 & y_1 & \cdots & y_t & \cdots & y_T
\end{bmatrix}$,
with $y_t\in\mathbb R^{n_y}$. 

\textit{Example:} Building thermal and refrigerant cycle dynamics are often represented by differential algebraic equations (DAEs) of the form
$0 = f_{\sf DAE}(\dot x, x, u, \theta_1)$ and
$y = h_{\sf DAE}(x, u, \theta_2)$.
One can  model this system using~\eqref{eq:model_map_mt} by considering $\theta:= \{\theta_1\}\cup\{\theta_2\}$ and simulating (i.e., numerically integrating) the system of DAEs forward over $t\in [0, T]$ to generate the sequence of outputs $y_{0:T}$.

The calibration task is to estimate a parameter set $\theta^\star\!\in\!\Theta$ that minimizes (in some goodness-of-fit sense) the modeling error 
$\ys - \mtstar$,
where $\ys$ denotes the measured outputs collected from a real system, and $\mtstar$ denotes the estimated outputs from the model $\mt$ using the estimated parameters $\theta^\star$. 
To this end, we propose optimizing a calibration cost function $J(\ys, \mt)$ to obtain the optimal parameters
\begin{align}\label{eq:cost}
\theta^\star &= \arg\min_{\theta\in\Theta} J(\ys, \mt).
\end{align}

Since each simulation is expensive and the underlying structure of the calibration model and cost are unknown, we solve the problem~\eqref{eq:cost} using Bayesian optimization (BO), which has shown potential for global optimization of black-box functions in a sample-efficient manner~\cite{snoek2012practical}. 
BO requires designing two components: a probabilistic map from the decision variables $\theta$ to the cost $J$, and an acquisition function that guides the selection of the next best optimizer candidate given the available data points. Classically, BO methods leverage Gaussian process (GP) regression for the task of providing a probabilistic map, but it is well known that GPs scale cubically with the number of available data points and the dimension of $\Theta$~\cite{snoek2015scalable}. Since we do not pose restrictions on $J$, it is possible that solving~\eqref{eq:cost} for large $n_\theta$ can  require thousands of data points to compute near-optimal solutions. This poses three critical challenges for classical BO methods: (C1) GP regression requires prohibitive training times with thousands of data points in high-dimensional spaces and are therefore not well-suited for calibrating large digital twins of modern buildings, (C2) the GP-approximated cost function is strongly dependent on the kernel selected by the user, and such kernels may induce functional properties like smoothness that are not always seen in practice; and, (C3) evaluating the cost function every time a new candidate parameter is computed is not amenable to parallelization. 

In lieu of GPs, we propose using attentive neural processes (ANPs) to approximate the calibration cost. ANPs are deep neural networks that are capable of learning a broad class of stochastic processes, and therefore, can make predictions equipped with uncertainty quantification~\cite{garnelo2018neural}. ANPs are highly scalable and suitable for training on high-dimensional problems with large datasets, and can perform well without requiring careful kernel selection. Consequently, we posit that replacing GPs with ANPs solves the challenges (C1) and (C2). Another benefit of ANPs is that they incur less computational complexity during inference than GPs. This fact, coupled with the observation that re-training an ANP with single datapoint increments seems wasteful as the inference of a deep neural network is unlikely to change significantly with one additional point, suggests the utility of batch BO (BBO) methods. Unlike BO, BBO acquisition functions generate a batch of candidates that are to be evaluated. Thus, the time-consuming cost function evaluation can be parallelized and the ANP updated with a batch of data points: this provides a way to address (C3). Note that due to multi-scale dynamics and combination of PDEs, DAEs, etc. in digital twins of buildings, simulating the twin can require orders of magnitude more time than retraining ANPs; especially simulations with large $T$.

\section{Relevant Work}
State-of-the-art methods for  calibration of  thermal models are presented in the survey by~\citet{wang2019data}: these models often do not consider equipment dynamics and parameters are not easily interpreted. Conversely, the study by~\citet{drgovna2021physics} shows the benefits of physics-informed models. However, the associated increase of digital twin model complexity requires scalable and sample-efficient optimization algorithms like BO.
\begin{figure*}[!ht]
	\centering
	\includegraphics[clip,width=\textwidth]{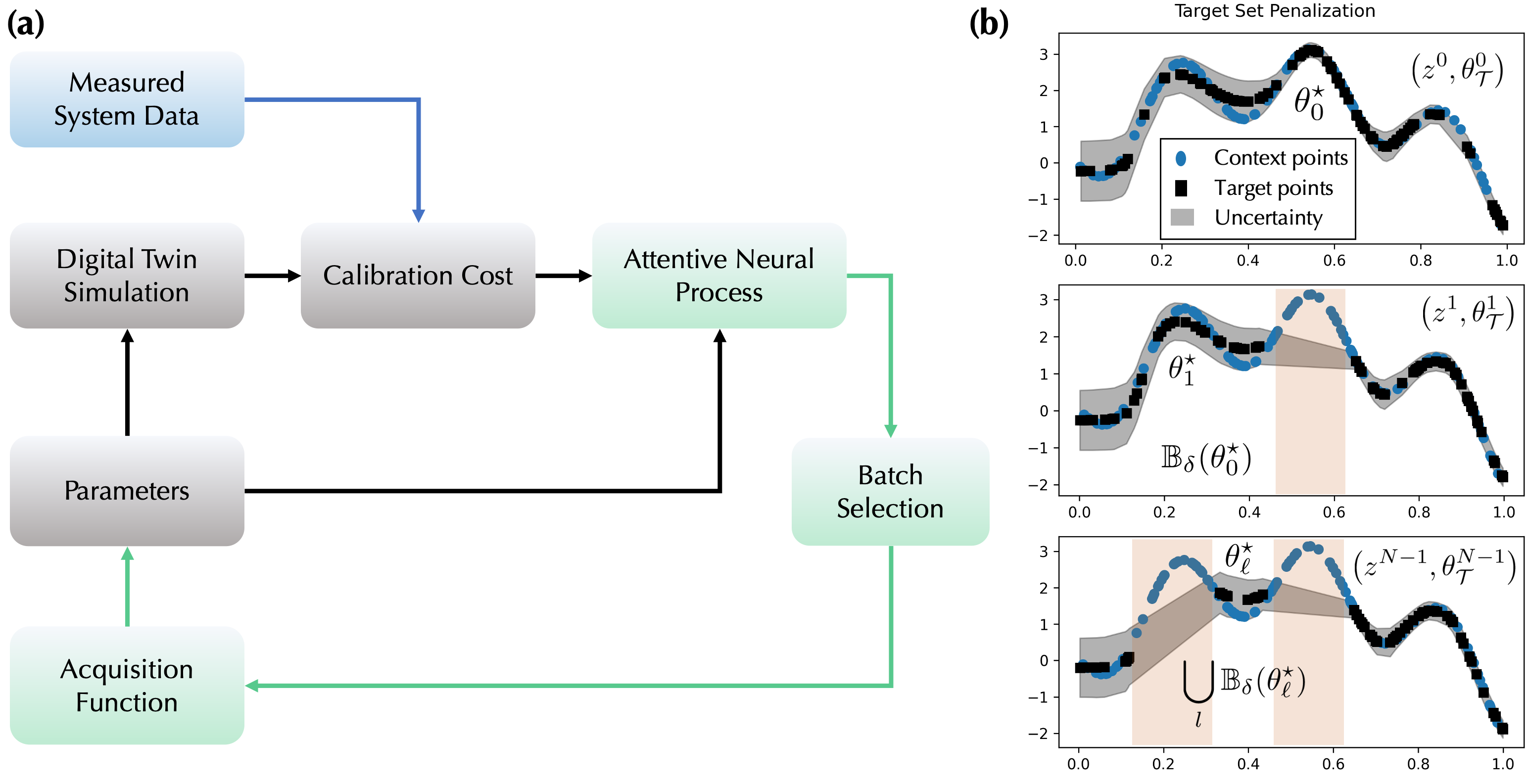}
	\vspace{-0.7em}
	\caption{(a) Calibration workflow with ANP-BBO. (b) Illustration of target penalization and latent sampling. More details about subplot (b) are provided in Appendix~\ref{sec:subplotB}.}
	\label{fig:ANP_BBO}
\end{figure*}
Scalable BO methods typically fall into two classes, those based on low dimensional embeddings~\cite{wang2016bayesian, nayebi2019framework, lu2018structured}, or those based on alternate probabilistic regressors that scale well with dimensions and number of data points, such as kernel methods~\cite{kandasamy2015high,oh2018bock}, or deep Bayesian
networks~\cite{snoek2015scalable,springenberg2016bayesian}. Very recently, a neural process (without attention) has been considered as a surrogate for BO~\cite{shangguan2021neural}. However, there is clear empirical evidence that ANPs produce `tight' predictions at data points where the objective is known; this is an essential property needed for BO and the lack of this property is a drawback of the NP. Furthermore, their implementation involves training the NP once with a large amount of initial data, which differs from our approach of batch BO to avoid re-training too often. 
Recent work on BBO has resulted in powerful algorithms for selecting batches based on the GP posterior~\cite{azimi2012hybrid, desautels2014parallelizing}, or by penalizing to find disjoint regions likely to contain extrema~\cite{gonzalez2016batch, nguyen2016budgeted}. Other BBO methods use hallucinations from the GP posterior either by Thompson sampling~\cite{depalma2019sampling} or multi-scale sampling of GP hyperparameters~\cite{joy2020batch}. In our proposed ANP-BBO, we combine the benefits of  penalization and  hallucination during ANP inference.

\section{The ANP-BBO Algorithm}

Fig.~\ref{fig:ANP_BBO} illustrates the ANP-BBO workflow for solving the optimization problem from~\eqref{eq:cost} in an iterative manner. Let $\mathcal D^t\!:=\!\{(\theta_{i}, J_{i})\}_{i=0}^{N_0 + tN}$ denote the data (parameter/cost pairs) collected up to the $t$-th iteration of ANP-BBO, where $N_0$ is the size of an initial dataset and $N$ is the batch-size, i.e., the number of model simulations/cost evaluations performed at each iteration. In this work, we use the ANP~\cite{kim2019attentive} to estimate the conditional Gaussian distribution $p(J_{\mathcal T}|\theta_\mathcal T,\mathcal D^t, z)$, where $J_{\mathcal T}$ is a set of cost function values  located at target points $\theta_\mathcal T\subset\Theta$, and $z$ is a latent variable that can be sampled to obtain different realizations of the learned stochastic process. A summary of relevant details of the ANP are provided in Appendix~\ref{ssec:anp}.

Directly substituting a GP with an ANP for BO with $N=1$ would involve: (i) training the ANP with $\mathcal D^t$ at every $t$, (ii) sampling target points $\theta_\mathcal T\subset\Theta$, (iii) obtaining one sample of the latent $z$, predict mean and variance of $J_\mathcal T$ for the target set, and evaluate an acquisition function at those points, (iv) selecting the target that maximizes the acquisition function as the best candidate $\theta^{t,\star}$, and (v) evaluating the cost for $\theta^{t,\star}$ and append this pair to $\mathcal D^t$, retrain the ANP, and repeat from (i) for iteration $t+1$. While not retraining the ANP is an option, this requires $N_0$ to be extremely large and the ANP trained with $N_0$ data points equipped with weights that reflect the underlying function closely. In our work, we follow the spirit of classical BO and assume $N_0$ is small. For this, the ANP needs retraining with a growing dataset. However, we posit that $N>1$ offers the significant advantages of reducing the number of times the ANP is retrained while enabling parallelized simulations of digital twins during evaluation of the calibration cost.

To this end, we propose the following modifications to the workflow (i)--(v) above; see Fig.~\ref{fig:ANP_BBO} for an illustration using a 1-D exemplar function. In particular, we loop through steps (ii)--(iv) $N$ times, and at each 
iteration $k=0,\cdots,N-1$, we perform target penalization by selecting a target set $\theta_\mathcal T^k$ away from neighborhoods of previous candidates. We explain the target penalization step formally as follows.
Let $\mathbb B_\delta(\theta)$ denote a ball of radius $\delta$ centered at $\theta\in\Theta$. 

At batch-selection iteration $k=0$, the target set is constructed by extracting samples from the entirety of $\Theta$. At each subsequent iteration, neighborhoods of $\Theta$ are removed to ensure diversity of solutions. Concretely, at the $k$-th iteration, if $\theta_{0:k-1}^{t,\star}$ denotes the set of candidates selected so far, then the target set $\theta_\mathcal T^k$ will be sampled from 
$
\Theta \setminus \left(\bigcup_{\ell=0}^{k-1}\mathbb B_\delta\big(\theta_{\ell}^{t,\star}\big)\right).
$
This method is similar to the local penalization approach of~\cite{gonzalez2016batch}, but we do not assume knowledge of Lipschitz constants of the cost. Our target penalization method rejects samples from the target set to ensure that candidates in the batch do not cluster around a suspected local minimum. To prevent conglomeration of candidates, we select $\delta$ large enough to maintain distance between candidates in the batch, while being small enough to ensure that fewer than $N$ balls cannot cover $\Theta$. 
Additionally, we utilize the ANPs ability to model families of distributions by sampling the latent variable iteratively during batch-selection. Sampling $z$ fixes a distribution from the family of distributions; thus, resampling $z$ during batch-selection promotes diversity in the statistics of the predicted output. At each $k$, the target sample that maximizes a given acquisition function is added to the batch. The target penalization and latent sampling can be highly parallelized for efficient cost function evaluation.

\section{Results and Discussion}

\textit{Setup:} Details about the physics-informed building digital twin are presented in Appendix~\ref{sec:digital-twin}. We obtain the data $y_{0:T}^\star$ by simulating the building dynamics for 5 days and collecting temperature and humidity measurements for 3 rooms (thus, $n_y=6$) every 15 minutes. The $n_\theta=12$ true parameters are provided in Table~\ref{tab:param_est_vals}. The measurements are corrupted by Gaussian noise of zero mean and 0.5 variance (for temperature) and 4 variance (for humidity); additionally, the sensors are assumed to be quantized at 0.1 resolution. The first $2$ days' data is used to train the ANP and perform calibration. The final 3 days are used for testing; that is, the calibrated digital twin predicts the outputs for the final 3 days for comparison with true outputs. 

\textit{Calibration performance:} Details about the ANP-BBO implementation are provided in Appendix~\ref{sec:anp_training}. Fig.~\ref{fig:building_outputs_comparison} illustrates the outputs over both training and testing data (5 days) after 1000 objective function evaluations. Clearly, the continuous lines (which are the digital twin predictions with the best parameters found by ANP-BBO) fit the data (colored circles) well, despite noise in measurements, for both temperature and humidity outputs. In fact, the coefficient-of-variation root-mean-squared error (CVRMSE: $\|\varepsilon_i\|/\sqrt{T}$) of all of our outputs are within 1\%, which is far below the ASHRAE guideline of 15\%~\cite{ASHRAE2014}. Furthermore, we are encouraged to see that despite considering a search space of significant volume (typically, calibration problems assume search space $\pm 20\%$ of the nominal parameter value), the final set of parameters are close to the true values (see Table~\ref{tab:param_est_vals}). Some estimates are better than others due to inherent sensitivities of the parameters, but most parameters are captured to over $90\%$ relative accuracy.

\begin{figure}[!ht]
	\centering
	\includegraphics[clip, width=\columnwidth]{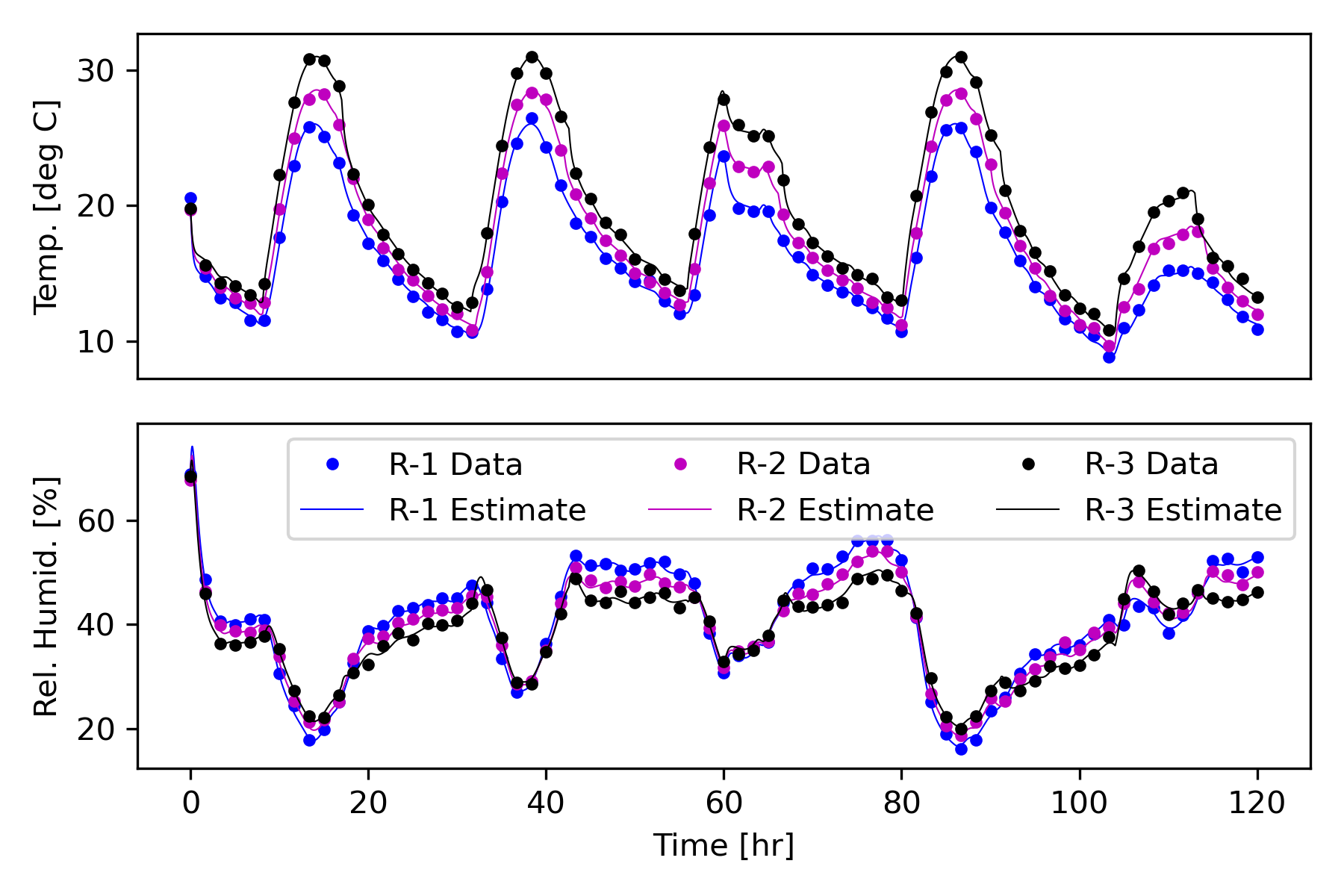}
	\caption{Test data and digital twin estimates for 5 days using the best set of parameters obtained by ANP-BBO. (R-x: Room x.)}
	\label{fig:building_outputs_comparison}
\end{figure}

\begin{table}[!ht]
	\scriptsize
    \centering
    \begin{tabular}{c|c|c|c||c|c|c|c}
    $\theta_i$ & True & Best & $\Theta_i$ & $\theta_i$ & True & Best & $\Theta_i$ \\\hline
    $\theta_1$ & 8.00 & 8.04 & [6, 10] &  $\theta_2$ & 5.00 & 5.09 &  [3, 7] \\
    $\theta_3$ & 0.45 & 0.35 & [0, 1] & $\theta_4$ & 3.00 & 2.88 & [2, 4] \\
    $\theta_5$ & 1.00 & 1.31 & [0, 2] & $\theta_6$ & 0.10 & 0.11 & [0, 1]\\
    $\theta_7$ & 1.00 & 0.97 & [0, 1] & $\theta_8$ & 0.10 & 0.07 & [0, 1]\\    
    $\theta_9$ & 18.00 & 18.35 & [14, 20] & $\theta_{10}$ & 10.00 & 10.32 & [8, 11] \\
    $\theta_{11}$ & 0.48 & 0.53 & [0,2] & $\theta_{12}$ & 6.00 & 6.07 & [3, 7]\\  \hline
    \end{tabular}
    \caption{Parameter estimates and corresponding search spaces.}    \label{tab:param_est_vals}
\end{table}

\begin{figure}[!ht]
	\centering
	\includegraphics[clip, width=\columnwidth]{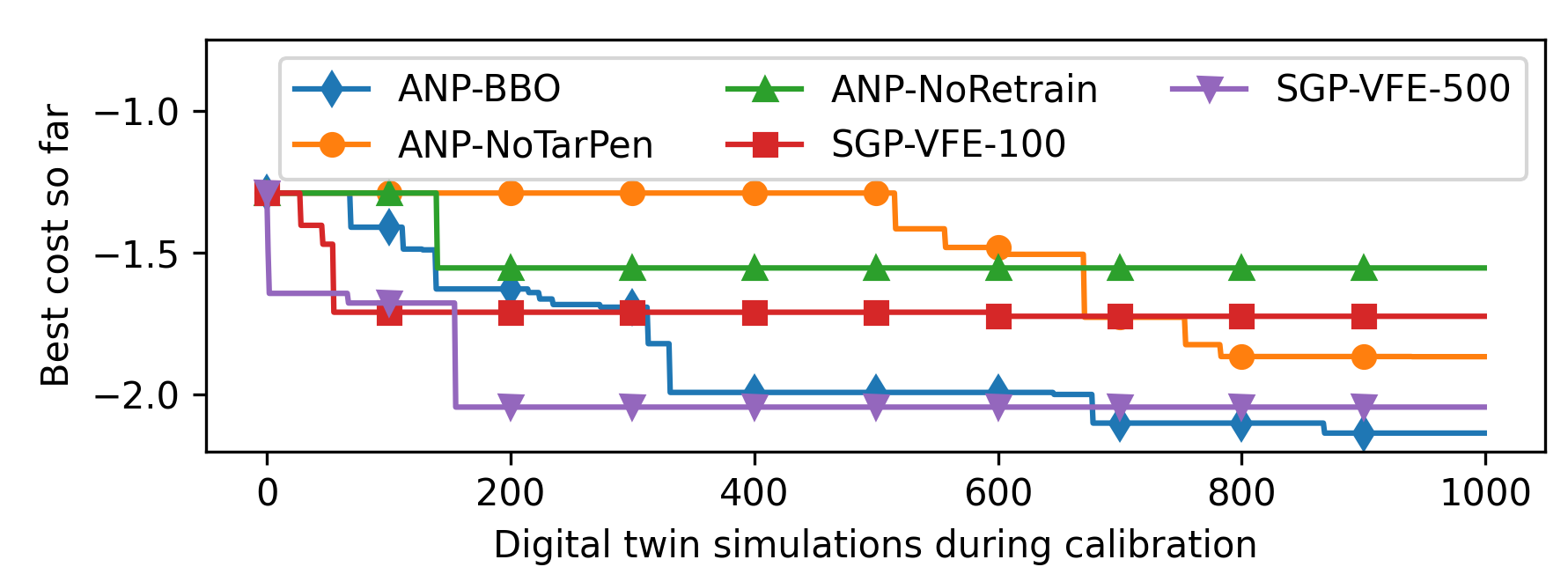}
	\caption{Ablation study results. Comparison of incumbent cost with number of function evaluations (i.e. number of simulations).}
	\label{fig:regretcomp}
\end{figure}

\textit{Ablation study:} We perform additional testing of ANP-BBO with the following modifications: (i) we switch off target set penalization and rely only on latent sampling for batch-selection ({ANP-NoTarPen}), (ii) we train the ANP once on the initial dataset and do not perform retraining ({ANP-NoRetrain}) as in~\citet{shangguan2021neural}, (iii) we perform BO with sparse Gaussian processes~\cite{titsias2009variational} with 100 inducing points and 1000 function evaluations (same as ANP-BBO) to prevent prohibitive training times ({SGP-VFE-100}), and (iv) same as (iii) but with 500 inducing points ({SGP-VFE-500}). The results of this study are encapsulated in Fig.~\ref{fig:regretcomp}, where we see that ANP-BBO outperforms its competitors, with SGP-VFE-500 showing fast decay but lack of improvement owing to subsequent BO candidates clustering around similar subregions of $\Theta$. The benefit of target penalization is also evident, as we see ANP-BBO's cost decays faster and more consistently than ANP-NoTarPen owing to the exploratory aspect introduced by the diversity amongst predictions induced by target penalization. We observe that ANP-NoRetrain performs  poorly, which is expected since lack of retraining implies that the attention weights are not recomputed, and therefore new context points which may contain critical information to the optimization problem is largely underutilized. 

To justify that ANP retraining is often faster than $N$ digital twin simulations, we refer the reader to Appendix~\ref{sec:timings}, where we have compared training and inference times of ANP and exact GP for a large number of datapoints in 12-D parameter space, as in our building calibration task. We also demonstrated that a week-long simulation of modern buildings that also accurately model stiff HVAC dynamics is comparable to ANP retraining times. Thus, we posit that longer horizon (month/year-long) or larger-scale (cities)  digital twin simulations will incur over $10\times$ wall-time than ANP retraining.

\section{Conclusions}
We proposed an ANP-BBO methodology that harnesses the power of probabilistic deep learning to calibrate industrial digital twins due to the presence of unmodeled dynamics and opacity incorporated to protect privacy, trade secrets, etc.
Precisely calibrating digital twins enables monitoring, control, self-optimization, and other key technologies that are strongly coupled with sustainability, air quality control, leakage detection, etc. Thus, \textit{accurate and scalable calibration mechanisms are essential to tackling climate change}. 

\bibliography{refs}
\bibliographystyle{tccmlicml2021}

\onecolumn

\section*{Appendix}
\renewcommand{\thesubsection}{\Alph{subsection}}

\subsection{Architecture and Description of Attentive Neural Processes (ANP)}
\label{ssec:anp}

In the context of Bayesian optimization for digital twin calibration, the ANP~\cite{kim2019attentive} is a regressor that defines stochastic processes with digital twin parameters serving as inputs $\theta_i\in\mathbb{R}^{n_\theta}$ , and function evaluations serving as outputs $J_i\in\mathbb{R}$. Given a dataset $\mathcal{D}=\{(\theta_{i}, J_{i})\}$, we learn an ANP for a set of $n_\mathcal T$ target points $\mathcal{D}_\mathcal{T}\subset \mathcal D$ conditioned on a set of $n_\mathcal C$ observed context points $\mathcal D_{\mathcal C}\subset \mathcal D$.  The ANP is invariant to the ordering of points in $\mathcal D_\mathcal T$ and $\mathcal D_\mathcal C$; furthermore, the context and target sets are not necessarily disjoint. The ANP additionally contains a global latent variable $z$ with prior $q(z|\mathcal D_{\mathcal C})$ that generates different stochastic process realizations, thereby incorporating uncertainty into the predictions of target function values $J_\mathcal T$ despite being provided a fixed context set.

Concretely, given a context set $\mathcal D_\mathcal C$ and target query points $\theta_\mathcal T$, the ANP estimates the conditional distribution of the target values $J_{\mathcal T}$ given by
\[
p(J_{\mathcal T}|\theta_\mathcal T,\mathcal D_\mathcal C):= \int p(J_{\mathcal T}|\theta_\mathcal T, r_\mathcal C, z)\,q(z|s_\mathcal C)\,\mathrm{d}z,
\]
where $r_\mathcal C:=r(\mathcal D_\mathcal C)$ is the output of the transformation induced by the \textit{deterministic} path of the ANP, obtained by aggregating the context set into a finite-dimensional representation that is invariant to the ordering of context set points (\textit{e.g.}, passing through a neural network and taking the mean). The function $s_\mathcal C:=s(\mathcal D_\mathcal C)$ is a similar permutation-invariant transformation made via a \textit{latent} path of the ANP. 
Both the transformations $r$ in the deterministic path and $s$ in the latent path are evaluated using self-attention networks~\cite{vaswani2017attention} with neural weights $\omega_r\neq \omega_s$ before aggregation. The aggregation operator in the latent path is typically the mean, whereas for the deterministic path, the ANP aggregates using a cross-attention mechanism, where each target query attends to the context points $\theta_\mathcal C$ to generate $r_{\mathcal C\times \mathcal T}(J_\mathcal T|\theta_\mathcal T, r_{\mathcal C}, z)$. 
Note that the ANP builds on the variational autoencoder (VAE) architecture, wherein $q(z|s)$, $r_\mathcal C$, and $s_\mathcal C$ form the encoder arm, and $p(J|\theta, r_{\mathcal C\times \mathcal T}, z)$ forms the decoder arm. The architecture of ANP with both paths is provided in Appendix-Fig.~\ref{fig:ANP_architecture}. 

For implementation, we make simplifying assumptions: (1) that each point in the target set is derived from conditionally independent Gaussian distributions, and (2) that the latent distribution is a multivariate Gaussian with a diagonal covariance matrix. This enables the use of the reparametrization trick~\cite{kingma2015variational} and train the ANP to maximize the evidence-lower bound loss
$$
\mathsf{E}\big[\log p(J_\mathcal T|\theta_\mathcal T, r_{\mathcal C\times\mathcal T}, z)\big] - \mathsf{KL}\left[q(z|s_\mathcal T)||q(z|s_\mathcal C)\right]
$$
for randomly selected $\mathcal D_\mathcal C$ and $\mathcal D_\mathcal T$ within $\mathcal D$. Maximizing the expectation term $\mathsf E(\cdot)$ ensured good fitting properties of the ANP to the given data, while minimizing (maximizing the negative of) the $\mathsf{KL}$ divergence embeds the intuition that the targets and contexts arise from the same family of stochastic processes. The complexity of ANP with both self-attention and cross-attention is $\mathbf O\left(n_\mathcal C(n_\mathcal C+n_\mathcal T)\right)$. Empirically, we observed that only using cross-attention does not deteriorate performance while resulting in a reduced complexity of approximately $\mathbf{O}\left(n_\mathcal C n_\mathcal T\right)$, which is beneficial because $n_\mathcal T$ is fixed, but $n_\mathcal C$ grows with BO iterations.

\begin{figure*}[!ht]
\includegraphics[clip,width=\textwidth]{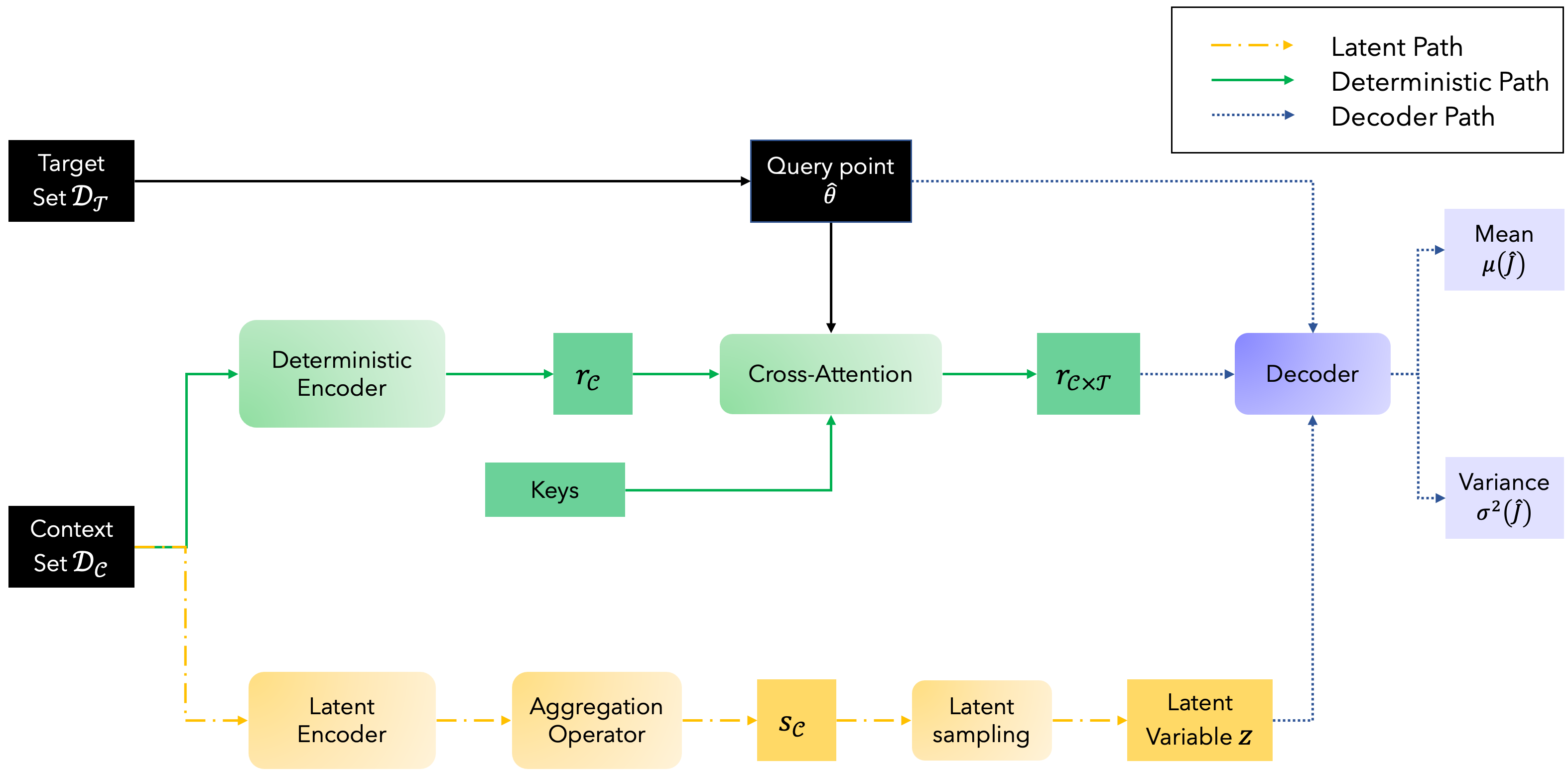}
\caption{ANP Architecture: Paths for training (latent, deterministic) and decoder.}
\label{fig:ANP_architecture}
\end{figure*}

\newpage
\subsection{More details about Fig. 1 subplot (b): }
\label{sec:subplotB}
The 1-D function used for illustration is given by $J(\theta):=\sin(20\theta) + \left(\tfrac{10\theta}{3}\right)^2 - 10\theta$ which has its global maximizer at $\theta^\star=0.5445$ on the search space $\Theta=[0, 1]$. The ANP for this example has been trained using $N_0=100$ initial samples randomly extracted from $\Theta$.  For each subplot in (b), the blue circles denote context points (these are the same for all subplots), the black squares are target points (these are penalized at subsequent batch-selection iterations after the first), the black shading denotes $\mu \pm 1.96\sigma$ around the ANP predictions at \textit{target} points, and the orange vertical shades depict subsets of $\Theta$ that are penalized (this is why there are no target points there). The top-most subplot of (b), with $k=0$, uses all target points to make a selection $\theta_0^\star=0.53$ for the batch. Since $\delta=0.1$, in the next batch-selection iteration $k=1$,   there are no target points in $\mathbb B_\delta(\theta_0^\star)$, which is why $\theta_1^\star=0.21$. The process is repeated. The effect of latent sampling is visible most clearly for $\theta\approx 0.8$, since the uncertainty bands there have larger lobes; empirically we have observed that this has a strong effect on the prediction uncertainties when only a few true data points have been obtained, which is the case in initial BO iterations.  

\newpage
\subsection{Description of Building Digital Twin}\label{sec:digital-twin}

\begin{figure*}[!ht]
	\includegraphics[clip,width=\textwidth]{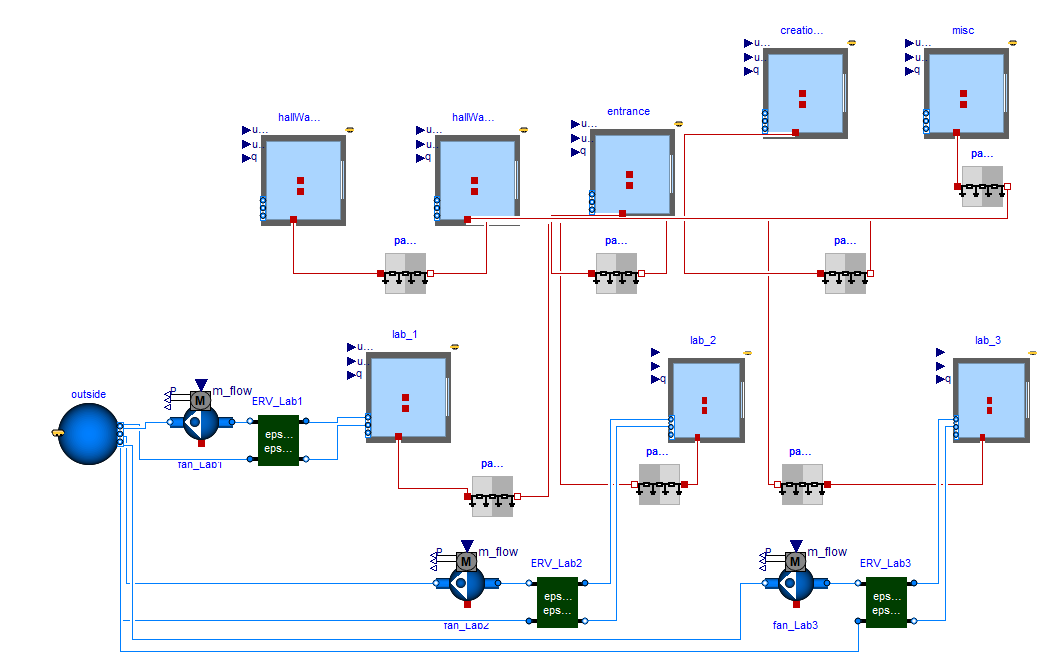}
	\caption{Building Plenum Component of Digital Twin: Modelica Implementation.}
	\label{fig:Plenum}
\end{figure*}

A model of one floor of a contemporary four story office building, located in the Tokyo, Japan geographic area, was used for this study.  We examine the behavior of three laboratory spaces (rooms) in this building, labeled \texttt{lab\_1}, \texttt{lab\_2}, and \texttt{lab\_3} in Figure~\ref{fig:Plenum}, over the course of 5 days of simulated behavior.  Each of these laboratory rooms has a floor area of 172.8 $\mathrm{m^2}$ and is 3.1 $\mathrm{m}$ tall, with one exterior facade that includes a single-pane 23 $\mathrm{m^2}$ window.  Conventional building materials and construction practices were used for the envelope, with an adiabatic lower boundary and a 1.15 m plenum above the laboratories with an adiabatic upper boundary above the plenum.  Other adjacent spaces were also included in the model, though the thermal interactions between the laboratories under study and these spaces were limited.  

This building model also included a ventilation system to supply fresh outside air for each of the laboratories.  These consisted of a fan providing inlet air at a flow rate of 2118 cfm, or approximately 0.5 air changes/hour.  This ventilation air was processed with an energy recovery ventilator (ERV) with a constant efficiency of 0.8 and pressure drop of 200~Pa to exchange thermal energy between the supply and exhaust air streams.   

Different occupancy schedules and loads were imposed on each of the rooms to explore the effect of different dynamics on the calibration process.  Laboratory 1 was occupied between the hours of 5am and 2pm, with a base convective/radiant load of 5 $\mathrm{W/m^2}$ and an occupied load of 14 $\mathrm{W/m^2}$, as well as 3 $\mathrm{W/m^2}$ of latent load during the occupied hours.  Laboratory 2 had a much lower overall load and was occupied between the hours of 8:30 am and 6pm, with a base convective/radiant load of 0.1 $\mathrm{W/m^2}$ and an occupied load of 1.1 $\mathrm{W/m^2}$, as well as  0.1 $\mathrm{W/m^2}$ of latent load during the occupied hours.  Finally, laboratory 3 had the highest overall load and was occupied between the hours of 3pm and 12am, with a base convective/radiant load of 10 $\mathrm{W/m^2}$ and an occupied load of 28 $\mathrm{W/m^2}$, as well as 6 $\mathrm{W/m^2}$ of latent load during the occupied hours.  

This model was constructed using the Modelica Buildings library~\cite{WetterMEB}, an open source library developed primarily by the Lawrence Berkeley National Laboratory to characterize a wide variety of components used in today's building systems.  These models characterize the convective, radiative, and latent heat transfer observed in occupied spaces, and employ an ideal gas moist air mixture model.  The Tokyo-Hyakuri TMY3 model was used to describe the weather conditions and solar heat gains between November 23-28 that were used as the subject of study, which is calibrated on recent climate data. 

\newpage
\subsection{Implementation Details}\label{sec:anp_training}

\subsubsection*{Codebase}
The ANP pipeline is implemented entirely in PyTorch~\cite{pytorch-ref}. 

Comparisons with GP are performed using GPyTorch~\cite{gpytorch-ref}.

\subsubsection*{ANP Implementation}
For the ANP model the architecture we follow the basic version from~\cite{kim2019attentive}, without self attention.  We use latent dimension of 128 for both the deterministic and latent paths.  The deterministic encoder uses 3 fully-connected layers with Leaky ReLU (slope=0.1) activations and 256 hidden units per layer.  The latent encoder uses a similar fully-connected architecture, however after the mean aggregation operation, we pass it through two fully connected layers, the first with linear activation functions, to obtain the latent mean $\mu(z)$, and the second with an activation of $0.1 +0.9\cdot \mathrm{sigmoid(x)}$ to obtain the latent standard deviation $\sigma(z)$.  For the cross attention block in the deterministic encoder, we first run the query and key coordinate positions through a fully-connected layer of size equal to the latent dimension to obtain learned positional encodings prior to the 8-head multi-head attention operation~\cite{vaswani2017attention}.  Finally the decoder which takes as input the concatenation of the sampled latent $z$, target position $\theta_{t_i}$, and the deterministic path target encoding $r_{t_i}$ consists of 3 fully-connected layers with Leaky ReLU (slope=0.1) activations and 256 hidden units per layer.  Finally, following the best practices in~\cite{le2018empirical}, the output layer in the decoder has two output units, the first with a linear activation function for estimating $\mu(J)$, and the second for estimating $\sigma(J)$ with a regularized softplus activation to avoid the standard deviation collapsing to zero, i.e., $0.1 +0.9\cdot \mathrm{softplus(x)}$.

We train the initial ANP for 5000 iterations (this is done offline with the $N_0$ initial data points) with a learning rate of $10^{-5}$, which is decreased by a factor of 2, after 1000 steps, and decreased by an additional factor of 5 after 2500 steps. We train the ANP at each BBO iteration $t$, using the ADAM optimizer with an initial learning rate of $5\times 10^{-5}$, which is decreased by a factor of 2, after 250 steps, and decreased by an additional factor of 5 after 500 steps.  Each step consists of a mini-batch of 32 randomly selected context ($\mathcal D_\mathcal C$) and target sets ($\mathcal D_\mathcal T$) within $\mathcal D$.  
For each element of the batch, we select points for $\mathcal D_\mathcal C$ and $\mathcal D_\mathcal T$ uniformly at random from all points in $\mathcal D$.  For all, BBO iterations after the initial one, we warm start the model using the weights from the previous iteration.

\subsubsection*{ANP-BBO Implementation}
We start with an initial set of $N_0=1000$ data points, where $\theta$ is sampled on $\Theta$ (provided in Table~\ref{tab:param_est_vals}) via Sobol sequences. The ANP is trained offline on this data, and the weights are stored for warm-starting subsequent retrains. With $\varepsilon(T):=(y_{0:T}^\star - \mathcal M_T(\theta))$, we select the cost $$J(\theta) = \log \left(\sum_{i=1}^{n_y}  \varepsilon_i(T)^\top W_i \varepsilon_i(T)\right),$$
where the logarithm helps with numerical conditioning and $W$ is chosen to scale the outputs to similar magnitudes. Note that we transform the minimization problem~\eqref{eq:cost} to a maximization problem, as in classical BO, by reversing the sign of $J$.
We then perform ANP-BBO for $N=200$ iterations with a batch-size of $K=5$ per iteration, an upper-confidence bound acquisition function $\mu + 3\sigma$, 5000 target points for sampling to obtain acquisition function maxima. For target penalization, we set $\delta=0.01$.

\newpage
\subsection{Comparison of Wall-times}
\label{sec:timings}

This comparison study was performed on a Windows 10 desktop with 32-GB RAM, Intel(R) Core(TM) i9-9900K CPU @ 3.60GHz. \underline{No GPU acceleration was used} for training either the ANP or GP methods. All training points are 12-dimensional vectors in accordance with the building digital twin parameters.

\begin{figure}[!ht]
	\centering
	\includegraphics[clip, width=\columnwidth]{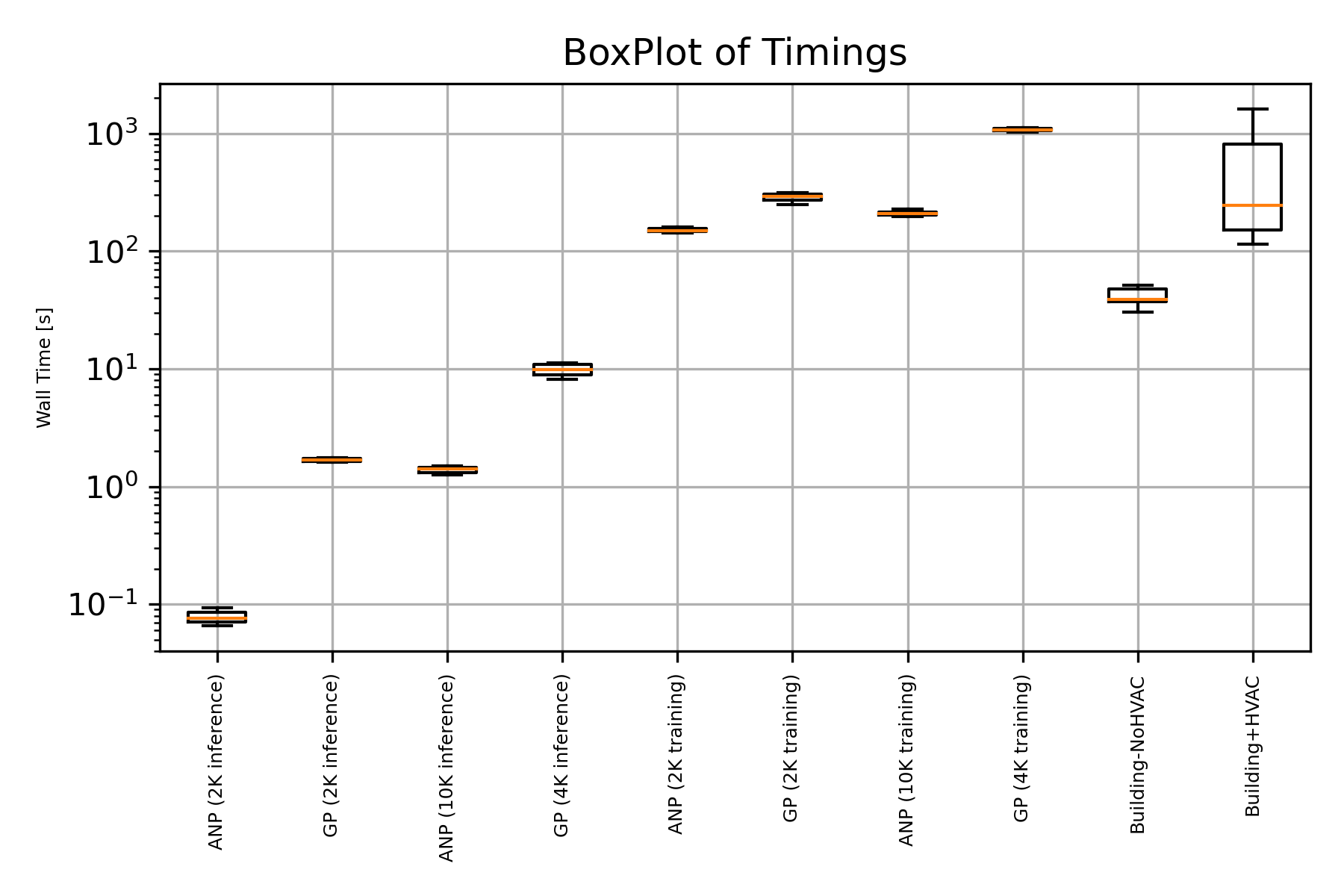}
	\caption{Wall times required for training, inference, and simulation of digital twin with and without HVAC equipment dynamics.}
	\label{fig:timings}
\end{figure}

While ANP training times can be large, we demonstrate that a digital twin simulation can be significantly larger depending on the time-span of simulation. In Fig.~\ref{fig:timings}, we compare the wall-time incurred on the same computer for 1000 training iterations for ANP with 2000 data points, ANP with 10,000 data points, GP with 2000 data points, GP with 4000 data points (after which GP becomes prohibitively slow). We also compare wall-times for inference of the same number of data points with ANP and GP. Finally, we present simulation times for 1 week using two digital twins: one with simple building dynamics and the cooling/heating system replaced by a lookup table, and another where the HVAC dynamical equations are also incorporated in the twin. Note that these twins could be simulated for month-long (or year-long) time-spans, which would require over 4$\times$ (or 52$\times$) these wall-times. Digital twins for cities and climate models would require considerably more wall-time to simulate, even with GPU integration; in fact, these Earth-scale simulations are often estimated to require $>5000$ GPUs~\cite{bauer2021digital} or supercomputing.
\end{document}